\documentclass[10pt,journal,compsoc]{IEEEtran}
\ifCLASSOPTIONcompsoc
  \usepackage[nocompress]{cite}
\else
  \usepackage{cite}
\fi

\ifCLASSINFOpdf
\else
\fi
\usepackage{graphicx} 
\usepackage{subfigure}
\usepackage{booktabs}
\usepackage{multirow}
\usepackage{amsmath} 
\usepackage{amssymb}           
\usepackage{amsfonts}            
\usepackage{mathrsfs}  
\usepackage{amsthm,amssymb}
\usepackage{color}
\usepackage{xcolor}
\usepackage{colortbl,booktabs}
\usepackage{fancybox}
\usepackage{framed} 
\usepackage{tikz}
\usepackage{pgfplots}
\usepackage{array}
\usepackage{microtype} 
\usepackage{tcolorbox} 
\usepackage[tikz]{bclogo}
\usepackage{lipsum} 
\usepackage{bbding}
\usepackage{pifont}
\usepackage{wasysym}
\usepackage{amssymb}
\usepackage[switch]{lineno}  %
\usepackage{pgfplots}
\pgfplotsset{width=7cm,compat=1.13}
\usepackage{subfigure}
\usepackage{url}
\usepackage{CJK}
\usepackage{ragged2e}

\begin{document}
%

\title{BURT: BERT-inspired Universal Representation\\ from Learning Meaningful Segment}
%
%
%
%

\author{Yian Li,
        Hai Zhao
\IEEEcompsocitemizethanks{\IEEEcompsocthanksitem Y. Li, H. Zhao are with the Department of Computer Science and Engineering, Shanghai Jiao Tong University, and also with Key Laboratory of Shanghai Education Commission for Intelligent Interaction and Cognitive Engineering, Shanghai Jiao Tong University, and also with MoE Key Lab of Artificial Intelligence, AI Institute, Shanghai Jiao Tong University.
\protect\\
E-mail: liya19@sjtu.edu.cn, zhaohai@cs.sjtu.edu.cn.
}
\IEEEcompsocitemizethanks{\IEEEcompsocthanksitem This paper was partially supported by National Key Research and Development Program of China (No. 2017YFB0304100), Key Projects of National Natural Science Foundation of China (U1836222 and 61733011), Huawei-SJTU long term AI project, Cutting-edge Machine Reading Comprehension and Language Model (Corresponding author: Hai Zhao).
}
\IEEEcompsocitemizethanks{\IEEEcompsocthanksitem This work has been submitted to the IEEE for possible publication. Copyright may be transferred without notice, after which this version may no longer be accessible.
}
}

%
%

\markboth{Journal of \LaTeX\ Class Files,~Vol.~14, No.~8, August~2015}%
{Shell \MakeLowercase{\textit{et al.}}: Bare Demo of IEEEtran.cls for Computer Society Journals}
%



\IEEEtitleabstractindextext{%
\begin{abstract}
\justifying
Although pre-trained contextualized language models such as BERT achieve significant performance on various downstream tasks, current language representation focuses on linguistic objective at a specific granularity. 
Thus this work introduces 
the universal representation learning, i.e., embeddings of different levels of linguistic unit in a uniform vector space. We present a universal representation model, \textbf{BURT} (\textbf{B}ERT-inspired \textbf{U}niversal \textbf{R}epresentation from learning meaningful segmen\textbf{T}), to encode different levels of linguistic unit into the same vector space. Specifically, we extract 
meaningful segments based on point-wise mutual information (PMI) to incorporate different granular objectives into the pre-training stage. We conduct experiments on datasets for English and Chinese including the GLUE and CLUE benchmarks, where our model surpasses its baselines and alternatives on a wide range of downstream tasks. We present our approach of constructing analogy datasets in terms of words, phrases and sentences and experiment with multiple representation models to examine geometric properties of the learned vector space through a task-independent evaluation. Finally, we verify the effectiveness of our method 
in two real-world text matching scenarios. As a result, our model significantly outperforms existing information retrieval (IR) methods and yields universal representations that can be directly applied to retrieval-based question-answering and natural language generation tasks. 
\end{abstract}

}

\maketitle

\IEEEdisplaynontitleabstractindextext

%
\IEEEpeerreviewmaketitle

\IEEEraisesectionheading{\section{Introduction}}
\IEEEPARstart{R}{epresentations} learned by deep neural models have attracted a lot of attention in Natural Language Processing (NLP). 
However, previous language representation learning methods such as Word2Vec \cite{word2vec}, LASER \cite{laser} and USE \cite{use} focus on either words or sentences. 
Later proposed pre-trained contextualized language representations like ELMo \cite{elmo}, GPT\cite{gpt-1}, BERT \cite{bert} and XLNet \cite{xlnet} may seemingly handle different sized input sentences, but all of them focus on sentence-level specific representation still for each word, leading to unsatisfactory performance in real-world situations. Although the latest BERT-wwm-ext \cite{bert-wwm}, StructBERT \cite{structbert} and SpanBERT \cite{spanbert} perform MLM on a higher linguistic level, the masked segments (whole words, trigrams, spans) either follow a pre-defined distribution or focus on a specific granularity. Besides, the random sampling strategy ignores important semantic and syntactic information of a sequence, resulting in a large number of meaningless segments.

However, universal representation among different levels of linguistic units may offer a great convenience when it is needed to handle free text in language hierarchy in a unified way. As well known that, embedding representation for a certain linguistic unit (i.e., word) enables linguistics-meaningful arithmetic calculation among different vectors, also known as word analogy. For example, \emph{vector (``King") - vector (``Man") + vector (``Woman")} results in \emph{vector (``Queen")}. Thus universal representation may generalize such good analogy features or meaningful arithmetic operation onto free text with all language levels involved together. For example, \textit{Eat an onion} : \textit{Vegetable} :: \textit{Eat a pear} : \textit{Fruit}. In fact, manipulating embeddings in the vector space reveals syntactic and semantic relations between the original sequences and this feature is indeed useful in true applications. For example, ``London is the capital of England.” can be formulized as $v(capital) + v(England) \approx v(London)$. Then given two documents one of which contains ``England” and ``capital”, the other contains ``London”, we consider these two documents relevant.

In this paper, we explore the regularities of representations including words, phrases and sentences in the same vector space. To this end, we introduce a universal analogy task derived from Google's word analogy dataset. To solve such task, we present BURT, a pre-trained model that aims at learning universal representations for sequences of various lengths. Our model follows the architecture of BERT but differs from its original masking and training scheme. Specifically, we propose to efficiently extract and prune meaningful segments (\textit{n}-grams) from unlabeled corpus with little human supervision, and then use them to modify the masking and training objective of BERT. The \textit{n}-gram pruning algorithm is based on point-wise mutual information (PMI) and automatically captures different levels of language information, which is critical to improving the model capability of handling multiple levels of linguistic objects in a unified way, i.e., embedding sequences of different lengths in the same vector space. 

Overall, our pre-trained models improves the performance of baselines in both English and Chinese. In English, BURT-base reaches 0.7 percent gain on average over Google BERT-base. In Chinese, BURT-wwm-ext obtains 74.5\% on the WSC test set, 13.4\% point absolute improvement compared with BERT-wwm-ext and exceeds the baselines by 0.2\% $\sim$ 0.6\% point accuracy on five other CLUE tasks including TNEWS, IFLYTEK, CSL, ChID and CMRC 2018. Extensive experimental results on our universal analogy task demonstrate that BURT is able to map sequences of variable lengths into a shared vector space where similar sequences are close to each other. Meanwhile, addition and subtraction of embeddings reflect semantic and syntactic connections between sequences. Moreover, BURT can be easily applied to real-world applications such as Frequently Asked Questions (FAQ) and Natural Language Generation (NLG) tasks, where it encodes words, sentences and paragraphs into the same embedding space and directly retrieves sequences that are semantically similar to the given query based on cosine similarity. All of the above experimental results demonstrate that our well-trained model leads to universal representation that can adapt to various tasks and applications.

\section{Background}
\subsection{Word and Sentence Embeddings}
Representing words as real-valued dense vectors is a core technique of deep learning in NLP. Word embedding models \cite{word2vec,glove,fasttext} map words into a vector space where similar words have similar latent representations. ELMo \cite{elmo} attempts to learn context-dependent word representations through a two-layer bi-directional LSTM network. In recent years, more and more researchers focus on learning sentence representations. The Skip-Thought model \cite{skip-thought} is designed to predict the surrounding sentences for an given sentence. Logeswaran and Lee (2018) \cite{quick-thought} 
improve the model structure by replacing the RNN decoder with a classifier. InferSent \cite{infersent} is trained on the Stanford Natural Language Inference (SNLI) dataset \cite{snli} in a supervised manner. Subramanian et al. (2018) \cite{gensen} and Cer et al. (2018) \cite{use} employ multi-task training and report considerable improvements on downstream tasks. LASER \cite{laser} is a BiLSTM encoder designed to learn multilingual sentence embeddings. Most recently, contextualized representations with a language model training objective such as OpenAI GPT \cite{gpt-1}, BERT \cite{bert}, XLNet \cite{xlnet} are expected to capture complex features (syntax and semantics) for sequences of any length. Especially, BERT improves the pre-training and fine-tuning scenario, obtaining new state-of-the-art results on multiple sentence-level tasks. On the basis of BERT, further fine-tuning using Siamese Network on NLI data can effectively produce high quality sentence embeddings \cite{sbert}. Nevertheless, most of the previous work concentrate on a specific granularity. In this work we extend the training goal to a unified level and enables the model to leverage different granular information, including, but not limited to, word, phrase or sentence.

\subsection{Pre-Training Tasks}
 BERT is trained on a large amount of unlabeled data including two training targets: Masked Language Model (MLM) for modeling deep bidirectional representations, and Next Sentence Prediction (NSP) for understanding the relationship between two sentences. ALBERT \cite{albert} is trained with Sentence-Order Prediction (SOP) as a substitution of NSP. StructBERT \cite{structbert} has a sentence structural objective that combines the random sampling strategy of NSP and continuous sampling as in SOP. However, RoBERTa \cite{roberta} and SpanBERT \cite{spanbert} use single contiguous sequences of 512 tokens for pre-training and show that removing the NSP objective improves the performance. Besides, BERT-wwm \cite{bert-wwm}, StructBERT \cite{spanbert}, SpanBERT \cite{structbert} perform MLM on higher linguistic levels, augmenting the MLM objective by masking whole words, trigrams or spans, respectively. Nevertheless, we concentrate on enhancing the masking and training procedures from a broader and more general perspective.

\subsection{Analysis and Applications}
Previous explorations of vector regularities mainly study word embeddings \cite{word2vec,phrase2vec}. After the introduction of sentence encoders and Transformer models \cite{transformer}, more works were done to investigate sentence-level embeddings. Usually the performance in downstream tasks is considered to be the measurement for model ability of representing sentences \cite{infersent,use,multidnn}. Some research proposes probing tasks to understand certain aspects of sentence embeddings \cite{probing_1,probing_2, probing_3}. Specifically, Rogers et al. (2020) \cite{bert_embedding_1} and Ma et al. (2019) \cite{bert_embedding_2} look into BERT embeddings and reveal its internal working mechanisms. Some work also explores the regularities in sentence embeddings \cite{linear_2, linear_3}. Nevertheless, little work analyzes words, phrases and sentences in the same vector space. In this paper, We work on embeddings for sequences of various lengths obtained by different models in a task-independent manner. 

Transformer-based representation models have made great progress in measuring query-Question or query-Answer similarities. Damani et al. (2020) \cite{faq_transformer} make an analysis on Transformer models and propose a neural architecture to solve the FAQ task. Sakata et al. (2019) \cite{faq_bert_1} come up with an FAQ retrieval system that combines the characteristics of BERT and rule-based methods. In this work, we also evaluate the well-trained universal representation models on FAQ task.

\begin{figure*}[t]
\centering
\includegraphics[width=0.8\textwidth]{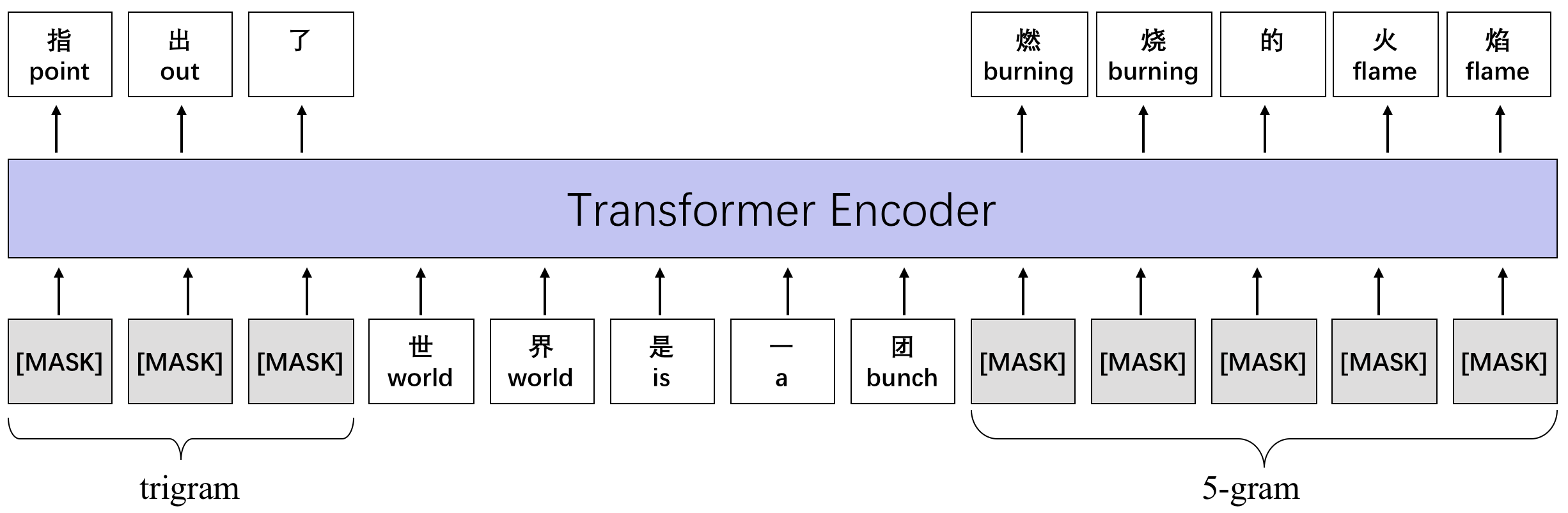}
\caption{An illustration of \textit{n}-gram pre-training.}
\label{model}
\end{figure*}

\begin{figure*}[t]
\centering
\includegraphics[width=0.8\textwidth]{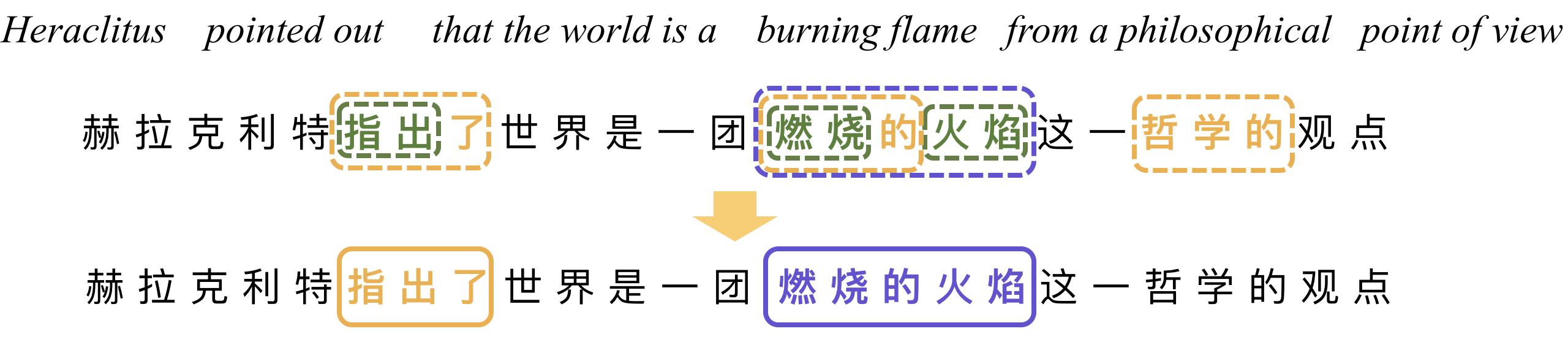}
\caption{An example from the Chinese Wikipedia corpus. \textit{n}-grams of different lengths are marked with dashed boxes in different colors in the upper part of the figure. During training, we randomly mask \textit{n}-grams and only the longest \textit{n}-gram is masked if there are multiple matches, as shown in the lower part of the figure.}
\label{ngram-example}
\end{figure*}

\section{Methodology}
Our BURT follows the Transformer encoder \cite{transformer} architecture where the input sequence is first split into subword tokens and a contextualized representation is learned for each token. We only perform MLM training on single sequences as suggested in \cite{spanbert}. The basic idea is to mask some of the tokens from the input and force the model to recover them from the context. Here we propose a unified masking method and training objective considering different grained linguistic units.

Specifically, we apply an pruning mechanism to collect meaningful \textit{n}-grams from the corpus and then perform \textit{n}-gram masking and predicting. Our model differs from the original BERT and other BERT-like models in several ways. First, instead of the token-level MLM of BERT, we incorporate different levels of linguistic units into the training objective in a comprehensive manner. Second, unlike SpanBERT and StructBERT which sample random spans or trigrams, our \textit{n}-gram sampling approach automatically discovers structures within any sequence and is not limited to any granularity.  

\subsection{\textit{N}-gram Pruning}
In this subsection, we introduce our approach of extracting a large number of meaningful $\emph{n}$-grams from the monolingual corpus, which is a critical step of data processing.

First, we scan the corpus and extract all $\emph{n}$-grams with lengths up to $N$ using the SRILM toolkit\footnote{http://www.speech.sri.com/projects/srilm/download.html} \cite{srilm}. In order to filter out meaningless $\emph{n}$-grams and prevent the vocabulary from being too large, we apply pruning by means of point-wise mutual information (PMI) \cite{pmi}. To be specific, mutual information $I(x, y)$ describes the association between tokens $x$ and $y$ by comparing the probability of observing $x$ and $y$ together with the probabilities of observing $x$ and $y$ independently. Higher mutual information indicates stronger association between the two tokens. 
\begin{equation}
I(x, y) = {\rm log} \frac{P(x, y)}{P(x) P(y)}    
\end{equation}
In practice, $P(x)$ and $P(y)$ denote the probabilities of $x$ and $y$, respectively, and $P(x, y)$ represents the joint probability of observing $x$ followed by $y$. This alleviates bias towards high-frequency words and allows tokens that are rarely used individually but often appear together such as ``\textit{San Francisco}" to have higher scores. In our application, an $\emph{n}$-gram denoted as $w = (x_1, \ldots, x_{L_w})$, where $L_w$ is the number of tokens in $w$, may contains more than two words. Therefore, we present an extended PMI formula displayed as below:
\begin{equation}
 PMI(w) = \frac{1}{L_w} \left( {\rm log}P(w) - \sum\limits_{k=1}^{L_w}{\rm log}P(x_k) \right)
\end{equation}
where the probabilities are estimated by counting the number of observations of each token and $\emph{n}$-gram in the corpus, and normalizing by the size of the corpus. $\frac{1}{L_w}$ is an additional normalization factor which avoids extremely low scores for longer \textit{n}-grams. Finally, $\emph{n}$-grams with PMI scores below the chosen threshold are filtered out, resulting in a vocabulary of meaningful \textit{n}-grams.

\subsection{\textit{N}-gram Masking}
For a given input $S = \{x_1, x_2, \dots, x_L\}$, where $L$ is the number of tokens in $S$, special tokens \texttt{[CLS]} and \texttt{[SEP]} are added at the beginning and end of the sequence, respectively. Before feeding the training data into the Transformer blocks, we identify all the \textit{n}-grams in the sequence using the aforementioned \textit{n}-gram vocabulary. An example is shown in Figure \ref{ngram-example}, where there are overlap between \textit{n}-grams, which indicates the multi-granular inner structure of the given sequence. In order to make better use of higher-level linguistic information, the longest \textit{n}-gram is retained if multiple matches exist. Compared with other masking strategies, our method has two advantages. First, \textit{n}-gram extracting and matching can be efficiently done in an unsupervised manner without introducing random noise. Second, by utilizing \textit{n}-grams of different lengths, we generalize the masking and training objective of BERT to a unified level where different granular linguistic units are integrated.

Following BERT, we mask 15\% of all tokens in each sequence. The data processing algorithm uniformly samples one \textit{n}-gram at a time until the maximum number of masking tokens is reached. 80\% of the time the we replace the entire \textit{n}-gram with \texttt{[MASK]} tokens. 10\% of the time it is replace with random tokens and 10\% of the time we keep it unchanged. The original token-level masking is retained and considered as a special case of \textit{n}-gram masking where $\textit{n}=1$. We employ dynamic masking as mentioned by Liu et al. (2019) \cite{roberta}, which means masking patterns for the same sequence in different epochs are probably different. 

\subsection{Traning Objective}
As depicted in Figure \ref{model}, the Transformer encoder generates a fixed-length contextualized representation at each input position and the model only predicts the masked tokens. Ideally, a universal representation model is able to capture features for multiple levels of linguistic units. Therefore, we extend the MLM training objective to a more general situation, where the model is trained to predict \textit{n}-grams rather than subwords.
\begin{equation}
    \max_{\theta} \sum_{w} {\rm log} P(w | \mathbf {\hat x}; \theta) = \max_{\theta} \sum_{(i, j)} {\rm log} P(x_i, \dots, x_j | \mathbf {\hat x}; \theta)
\end{equation}
where $w$ is a masked \textit{n}-gram and $ \mathbf {\hat x}$ is a corrupted version of the input sequence. $(i, j)$ represents the absolute start and end positions of $w$.

\begin{table}[t]{}
  \caption{Statistics of datasets from GLUE and CLUE benchmarks. \#Train, \#Dev, \#Test are the size of training, development and test sets, respectively. \#L is the number of labels. Sequences are simply divided into two categories according to their length: ``Long" and ``Short".}
  \label{clue-statistics}
  \centering
  \begin{tabular}{llrrrr}\toprule
  
  \textbf{GLUE} & Length &\#Train & \#Dev & \#Test & \#L \\\midrule
    CoLA		&Short&8.5k	&1k&1k&2 \\
    SST-2	&Short&	67k&872	&1.8k&2\\
    MNLI & Short-Short & 393k&20k&20k& 3\\
    QNLI	&Short-Short	& 105k	&5.5k&5.5k&2 \\
    RTE	&Short-Short	&2.5k&277&3k&2	\\
    MRPC &Short-Short	&3.7k&408&1.7k&	2\\
    QQP &Short-Short	&364k&40k&391k&	2\\
    STS-B &Short-Short	&5.8k&1.5k&1.4k&	-\\
    \midrule\midrule
    
    \textbf{CLUE} & Length &\#Train & \#Dev & \#Test & \#L \\\midrule
    TNEWS		&Short&53k	&10k&10k&15 \\
    IFLYTEK	&Long&	12k&	2.6k&2.6k&119\\
    WSC & Short & 1.2k&304&290& 2\\
    AFQMC	&Short-Short	& 34k	&4.3k&3.9k&2 \\
    CSL	&Long-Short	&20k&3k&3k&2	\\
    OCNLI &Short-Short	&50k&3k&3k&	3\\
    CMRC18 &Long &10k&1k &3.2k &-\\
    ChID & Long & 85k & 3.2k & 3.2k & -\\
    C\textsuperscript{3} &Short & 12k & 3.8k& 3.9k & -\\\bottomrule
  \end{tabular}
\end{table}

\section{Task Setup}
To evaluate the model ability of handling different linguistic units, we apply our model on downstream tasks from GLUE and CLUE benchmark. Moreover, we construct a universal analogy task based on Google's word analogy dataset to explore the regularity of universal representation. Finally, we present an insurance FAQ task and a retrieval-based language generation task, where the key is to embed sequences of different lengths in the same vector space and retrieve sequences with similar meaning to the given query.

\subsection{General Language Understanding}
Statistics of the GLUE and CLUE benchmarks are listed in Table \ref{clue-statistics}. Besides the diversity of task types, we also find that different datasets concentrates on sequences of different lengths, which satisfies our need to examine the model ability of representing multiple granular linguistic units. 

\subsubsection{GLUE}
The General Language Understanding Evaluation (GLUE) benchmark \cite{glue} is a collection of tasks that is widely used to evaluate the performance of English language models. We divide eight NLU tasks from the GLUE benchmark into three main categories.

\noindent
\textbf{Single-Sentence Classification}
The Corpus of Linguistic Acceptability (CoLA) \cite{cola} is to determine whether a sentence is grammatically acceptability or not. The Stanford Sentiment Treebank (SST-2) \cite{sst-2} is a sentiment classification task that requires the model to predict whether the sentiment of a sentence is positive or negative. In both datasets, each example is a sequence of words annotated with a label.

\noindent
\textbf{Natural Language Inference} Multi-Genre Natural Language Inference (MNLI) \cite{mnli}, Stanford Question Answering Dataset (QNIL) \cite{qnli} and Recognizing Textual Entailment (RTE) \cite{rte} are natural language inference tasks, where a pair of sentences are given and the model is trained to identify the relationship between the two sentences from \textit{entailment}, \textit{contradiction}, and \textit{neutral}.

\noindent
\textbf{Semantic Similarity} Semantic similarity tasks identify whether the two sentences are equivalent or measure the degree of semantic similarity of two sentences according to their representations. Microsoft Paraphrase corpus (MRPC) \cite{mrpc} and Quora Question Pairs (QQP) dataset are paraphrase datasets, where each example consists of two sentences and a label of ``1" indicating they are paraphrases or ``0" otherwise. The goal of Semantic Textual Similarity benchmark (STS-B) \cite{sts-b} is to predict a continuous scores from 1 to 5 for each pair as the similarity of the two sentences.

\subsubsection{CLUE} The Chinese General Language Understanding Evaluation (ChineseGLUE or CLUE) benchmark \cite{clue} is a Chinese version of the GLUE benchmark for language understanding. We also find nine tasks from the CLUE benchmark can be classified into three groups. 

\noindent
\textbf{Single Sentence Tasks} We utilize three single-sentence classification tasks including TouTiao Text Classification for News Titles (TNEWS), IFLYTEK \cite{iflytek} and the Chinese Winograd Schema Challenge (WSC) dataset. Examples from TNEWS and IFLYTEK are  short and long sequences, respectively, and the goal is to predict the category that the given single sequence belongs to. WSC is a coreference resolution task where the model is required to decide whether two spans refer to the same entity in the original sequence.

\noindent
\textbf{Sentence Pair Tasks}
The Ant Financial Question Matching Corpus (AFQMC), Chinese Scientific Literature (CSL) dataset and Original Chinese Natural Language Inference (OCNLI) \cite{ocnli} are three pairwise textual classification tasks. AFQMC contains sentence pairs and binary labels, and the model is asked to examine whether two sentences are semantically similar. Each example in CSL involves a text and several keywords. The model needs to determine whether these keywords are true labels of the text. OCNLI is a natural language inference task following the same collection procedures of MNLI.

\noindent
\textbf{Machine Reading Comprehension Tasks}
CMRC 2018 \cite{cmrc2018}, ChID \cite{chid}, and C\textsuperscript{3} \cite{c3} are span-extraction based, cloze style and free-form multiple-choice machine reading comprehension datasets, respectively. Answers to the questions in CMRC 2018 are spans extracted from the given passages. ChID is a collection of passages with blanks and corresponding candidates for the model to decide the most suitable option. C\textsuperscript{3} is similar to RACE and DREAM, where the model has to choose the correct answer from several candidate options based on a text and a question.

\begin{table}[t]
  \caption{Examples from our word analogy dataset. The correct answers are in bold.}
  \label{analogy_example}
  \centering
  \begin{tabular}{|p{3cm}|p{4cm}|}\hline
     A : B :: C & Candidates\\\hline
    \textit{boy}:\textit{girl}::\textit{brother} & \textit{daughter, \textbf{sister}, wife, father, son}\\\hline
    
     \textit{bad}:\textit{worse}::\textit{big} & \textit{\textbf{bigger}, larger, smaller, biggest, better}\\\hline
     
     \textit{Beijing}:\textit{China}::\textit{Paris} & \textit{\textbf{France}, Europe, Germany, Belgium, London}\\\hline
     
     \textit{Chile}:\textit{Chilean}::\textit{China} & \textit{Japanese, \textbf{Chinese}, Russian, Korean, Ukrainian}\\\hline
  \end{tabular}
\end{table}

\subsection{Universal Analogy}
As a new task, universal representation has to be evaluated in a multiple-granular analogy dataset. The purpose of proposing a task-independent dataset is to avoid determining the quality of the learned vectors and interpret the model based on a specific problem or situation. Since embeddings are essentially dense vectors, it is natural to apply mathematical operations on them. In this subsection, we introduce the procedure of constructing different levels of analogy datasets based on Google's word analogy dataset. 

\subsubsection{Word-level analogy}
Recall that in a word analogy task \cite{word2vec}, two pairs of words that share the same type of relationship, denoted as $A$ : $B$ :: $C$ : $D$, are involved. The goal is to solve questions like ``$A$ is to $B$ as $C$ is to ?", which is to retrieve the last word from the vocabulary given the first three words. The objective can be formulated as maximizing the cosine similarity between the target word embedding and the linear combination of the given vectors:
\begin{align*}
     &d^* = \mathop{\arg\max}_{d^*} cosine(c+b-a, d)\\
     &cosine(u, v) = \frac{u\cdot v}{\|u\|\|v\|}
\end{align*}
where $a$, $b$, $c$, $d$ represent embeddings of the corresponding words and are all normalized to unit lengths.

To facilitate comparison between models with different vocabularies, we construct a closed-vocabulary analogy task based on Google's word analogy dataset through negative sampling. Concretely, for each question, we use GloVe to rank every word in the vocabulary and the top 5 results are considered to be candidate words. If GloVe fails to retrieve the correct answer, we manually add it to make sure it is included in the candidates. During evaluation, the model is expected to select the correct answer from 5 candidate words. Examples are listed in Table \ref{analogy_example}. 

\subsubsection{Phrase/Sentence-level analogy}
To investigate the arithmetic properties of vectors for higher levels of linguistic units, we present phrase and sentence analogy tasks based on the proposed word analogy dataset. We only consider a subset of the original analogy task because we find that for some categories, such as ``Australia" : `` Australian", the same template phrase/sentence cannot be applied on both words. Statistics are shown in Table \ref{statistics}.

\begin{table}[t]{}
\caption{Statistics of our analogy datasets. \#p and \#q are the number of pairs and questions for each category. \#c is the number of candidates for each dataset. \#l (p/s) is the average sequence length in phrase/sentence-level analogy datasets.}
  \label{statistics}
  \centering
  \begin{tabular}{lrrrr}\toprule
    Dataset & \#p & \#q & \#c & \#l (p/s)\\\midrule
    capital-common	& 23	& 506	&5 &6.0/12.0\\
    capital-world	&116	&4524	&5 &6.0/12.0 \\
    city-state	&67&	2467&	5 &6.0/12.0\\
    male-female	&23	&506&	5 &4.1/10.1\\
    present-participle	&33	&1056&	2 &4.8/8.8\\
    positive-comparative	&37	&1322&	2 &3.4/6.1\\
    positive-negative &	29	&812	&2& 4.4/9.2\\\midrule
    All	&328	&11193	&- & 5.4/10.7\\\bottomrule
  \end{tabular}
\end{table}

\begin{figure*}[t]
\centering
\includegraphics[width=0.8\textwidth]{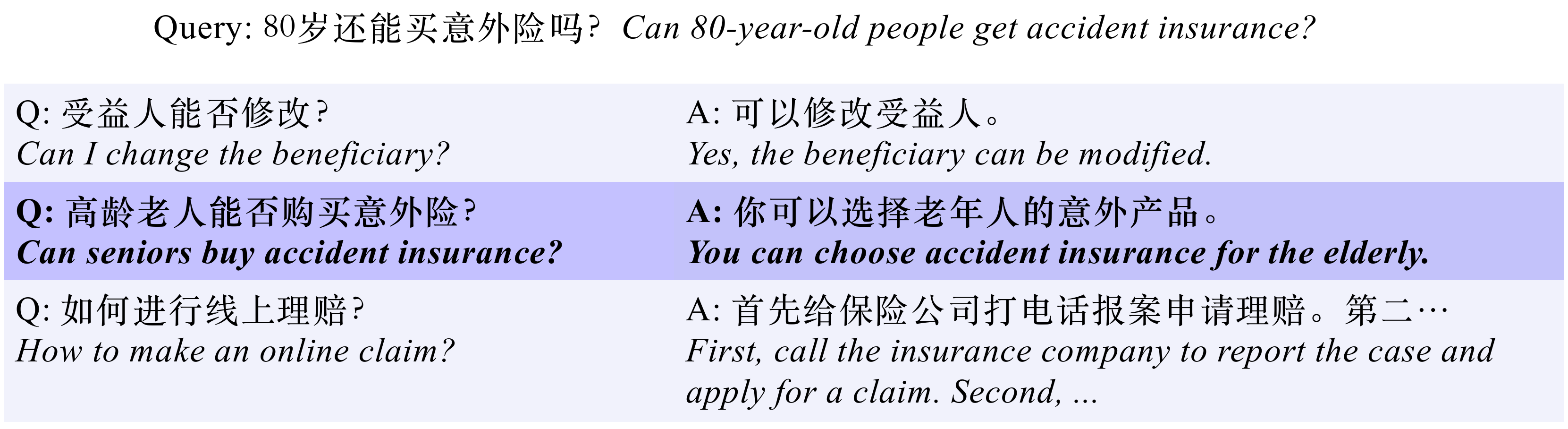}
\caption{Examples of Question-Answer pairs from our insurance FAQ dataset. The correct match to the query is highlighted.}
\label{faq-example}
\end{figure*}

\noindent
\textbf{Semantic}
Semantic analogies can be divided into four subsets: ``capital-common", ``capital-world", ``city-state" and ``male-female". The first two sets can be merged into a larger dataset: ``capital-country", which contains pairs of countries and their capital cities; the third involves states and their cities; the last one contains pairs with gender relations. Considering GloVe's poor performance on word-level ``country-currency" questions ($<$32\%), we discard this subset in phrase and sentence-level analogies. Then we put words into contexts so that the resulting phrases and sentences also have linear relationships. For example, based on relationship \textit{Athens} : \textit{Greece} :: \textit{Baghdad} : \textit{Iraq}, we select phrases and sentences that contain the word ``\textit{Athens}" from the English Wikipedia Corpus\footnote{https://dumps.wikimedia.org/enwiki/latest}: ``\textit{He was hired by the university of Athens as being professor of physics.}" and create examples: ``\textit{hired by ... Athens}" : ``\textit{hired by ... Greece}" :: ``\textit{hired by ... Baghdad}" : ``\textit{hired by ... Iraq}". However, we found that such a question is identical to word-level analogy for BOW methods like averaging GloVe vectors, because they treat embeddings independently despite the content and word order. To avoid lexical overlap between sequences, we replace certain words and phrases with their synonyms and paraphrases, e.g., ``\textit{hired by ... Athens}" : ``\textit{employed by ... Greece}" :: ``\textit{employed by ... Baghdad}" : ``\textit{hired by ... Iraq}".  Usually sentences selected from the corpus have a lot of redundant information. To ensure consistency, we manually modify some words during the construction of templates. However, this procedure will not affect the relationship between sentences.

\begin{table}[t]{}
  \caption{Details of the templates.}
  \label{nlg-statistics}
  \centering
  \begin{tabular}{lp{5cm}l}\toprule
    \textbf{Category} & Topics & \\\midrule
    Daily Scenarios	& Traveling, Recipe, Skin care, Beauty makeup, Pets& \;\;22 \\
    Sport \& Health	& Outdoor sports, Athletics, Weight loss, Medical treatment & \;\;15 \\
    Reviews 	& Movies, Music, Poetry, Books & \;\;16 \\
    Persons	& Entrepreneurs, Historical/Public figures, Writers, Directors, Actors& \;\;17 \\
    General&Festivals, Hot topics, TV shows & \;\;\;\;6 \\
    Specialized & Management, Marketing, Commerce, Workplace skills & \;\;17 \\
    Others	& Relationships, Technology, Education, Literature & \;\;14 \\\midrule
    All & -& 107 \\\bottomrule
  \end{tabular}
\end{table}

\noindent
\textbf{Syntactic}
We consider three typical syntactic analogies: Tense, Comparative and Negation, corresponding to three subsets: ``present-participle", ``positive-comparative", ``positive-negative", where the model needs to distinguish the correct answer from ``past tense", ``superlative" and ``positive", respectively. For example, given phrases ``\textit{Pigs are bright}" : ``\textit{Pigs are brighter than goats}" :: ``\textit{The train is slow}", the model need to give higher similarity score to the sentence that contains ``\textit{slower}" than the one that contains ``\textit{slowest}". Similarly, we add synonyms and synonymous phrases for each question to evaluate the model ability of learning context-aware embeddings rather than interpreting each word in the question independently. For instance, ``\textit{pleasant}" $\approx$ ``\textit{not unpleasant}" and ``\textit{unpleasant}" $\approx$ ``\textit{not pleasant}".

\subsection{Retrieval-based FAQ}
The sentence-level analogy discovers relationships between sentences by directly manipulating sentence vectors. Especially, we observe that sentences with similar meanings are close to each other in the vector space, which we find is consistent with the target of information retrieval task such as Frequently Asked Question (FAQ). Such task is to retrieve relevant documents ( FAQs) given a user query, which can be accurately done by only manipulating vectors representing the sentences, such as calculating and ranking vector distance in terms of cosine similarity. Thus, we present an insurance FAQ task in this subsection to explore the effectiveness of BURT in real-world retrieval applications.

An FAQ task involves a collection of Question-Answer (QA) pairs denoted as $\{(Q_1, A_1), (Q_2, A_2), ... (Q_N, A_N)\}$, where $N$ is the number of QA pairs. The goal is to retrieve the most relevant QA pairs for a given query. We collect frequently asked questions and answers between users and customer service from our partners in a Chinese online financial education institution. It contains over 4 types of insurance questions, e.g., concept explanation (``\textit{what}"), insurance consultation (``\textit{why}", ``\textit{how}"), judgement (``\textit{whether}") and recommendation. An example is shown in Figure \ref{faq-example}. Our dataset is composed of 300 QA pairs that are carefully selected to avoid similar questions so that each query has only one exact match. Because queries are mainly paraphrases of the standard questions, we use query-Question similarity as the ranking score. The test set consists of 875 queries and the average lengths of questions and queries are 14 and 16, respectively. The evaluation metric is Top-1 Accuracy (Acc.) and Mean Reciprocal Rank (MRR) because there is only one correct answer for each query. 
\begin{table*}[t]{}
  \caption{CLUE test results scored by the evaluation server\protect\footnotemark. ``acc" and ``EM" denote accuracy and Exact Match, respectively.}
  \label{clue-results}
  \centering
  \begin{tabular}{llcccccccccccc}
  
  \multicolumn{13}{c}{\textit{Batch size: 8, 16; Length: 128, 256; Epoch: 2, 3, 5, 50; lr: 1e-5, 2e-5, 3e-5}}\\ \toprule
  
  \multirow{3}{*}{\textbf{Models}}
  & \multicolumn{3}{c}{\textbf{Single Sentence}} && \multicolumn{3}{c}{\textbf{Sentence Pair}} &&
  \multicolumn{3}{c}{\textbf{MRC}}&\multirow{3}{*}{\textbf{Avg.}}\\\cline{2-4} \cline{6-8} \cline{10-12}
  \rule{0pt}{12pt}
    & TNEWS & IFLYTEK & WSC && AFQMC & CSL & OCNLI && CMRC18 & ChID & C\textsuperscript{3}& \\
    & (acc) & (acc)& (acc) && (acc) & (acc) & (acc) && (EM) & (acc) &(acc) &
     \\\midrule
    
    BERT 	&56.6 &60.3 &62.0&&73.7 &80.4&72.2&&71.6& 82.0& 64.5 & 69.3\\
    MLM &56.5 &60.2 &70.7&& 73.3 &79.3&70.6&&69.1 & 81.3& 64.8 &69.5 \\
    Span &56.7&59.6&72.1&&73.5&79.7&71.0&&71.6 & 82.2& 65.3 & 70.2\\
  \textbf{BURT}	&56.9 &\textbf{60.5}&74.1&&73.1  &80.8&71.3&& 71.7 & 82.2 & 65.7 &70.7\\
                \midrule
    BERT-wwm-ext  &56.8 &59.4 &61.1&&\textbf{74.1}  &80.6&\textbf{73.4}&&74.0 & 82.9& \textbf{68.5} & 70.1\\
    MLM&56.7&59.4&71.0&&74.0&80.4&72.8&&73.1 &  82.1 & 67.4 & 70.8\\
    Span &56.9&58.5&73.8&& 73.2&80.2&71.6&&72.4 & 82.2 & 67.1 & 70.7\\
    \textbf{BURT-wwm-ext}	 &\textbf{57.3} &60.1&\textbf{74.5} &&73.8&\textbf{81.0} &72.2 &&\textbf{74.2}  & \textbf{83.0 }& 67.6 & \textbf{71.5}\\\bottomrule
  \end{tabular}
\end{table*}
\footnotetext{https://www.cluebenchmarks.com/rc.html}

\begin{table*}[t]{}
  \caption{GLUE test results scored by the evaluation server\protect\footnotemark. We exclude the problematic WNLI set and recalculate the ``Avg." score. Results for BERT-base and BERT-large are obtained from \cite{bert}. ``mc" and ``pc" are Matthews correlation coefficient \cite{mc} and Pearson correlation coefficient, respectively.}
  \label{clue-results}
  \centering
  \begin{tabular}{llccccccccccccc}
  
  \multicolumn{12}{c}{\textit{Batch size: 8, 16, 32, 64; Length: 128; Epoch: 3; lr: 3e-5}}\\ \toprule
  
  \multirow{3}{*}{\textbf{Models}}
  & \multicolumn{2}{c}{\textbf{Single Sentence}} && \multicolumn{3}{c}{\textbf{NLI}} &&
  \multicolumn{3}{c}{\textbf{Semantic Similarity}}&\multirow{3}{*}{\textbf{Avg.}}\\\cline{2-3} \cline{5-7} \cline{9-11}
  \rule{0pt}{12pt}
    & CoLA & SST-2&& MNLI & QNLI & RTE && MRPC & QQP & STS-B & \\
    & (mc) & (acc)&& m/mm(acc) & (acc) & (acc) && (F1) & (F1) & (pc)  &
     \\\midrule
    
    BERT-base 	&52.1 &93.5 && 84.6/83.4 &90.5&	66.4 && 88.9	&	71.2& 87.1& 79.7\\
    MLM &51.9&	93.5&& 84.5/83.9&	90.7&	65.0&& 88.1	&	71.6& 86.2 & 79.5\\
    Span &53.3&	93.8&&84.5/84.0	&90.9&	66.6&& 88.0	&	71.6& 86.1& 79.9 \\
  \textbf{BURT-base}	& 55.7	&94.5&&84.7/84.1&	91.1&	67.1&&88.2	&	71.6 &86.4 & 80.4\\
                \midrule
    BERT-large  & 60.5	&\textbf{94.9} &&86.7/85.9 &\textbf{92.7}&	70.1&& 89.3	&	72.1& \textbf{87.6}& 82.2\\
    MLM&61.1&	94.5 && 86.6/85.6&	92.5&	69.2 &&\textbf{90.2}	&	\textbf{72.3}& 87.0& 82.1\\
    Span &60.1&	94.8&& 86.5/85.9&	92.6&	69.4&& 89.3	&	\textbf{72.3}& 87.3 & 82.0\\
    \textbf{BURT-large} &\textbf{62.6}&	94.7 && \textbf{86.8}/\textbf{86.0} &\textbf{92.7}&	\textbf{70.8}&& 89.7	&	\textbf{72.3} &87.3& \textbf{82.5}\\\bottomrule
  \end{tabular}
\end{table*}
\footnotetext{https://gluebenchmark.com}

\subsection{Natural Language Generation}
Moving from word and sentence vectors towards representation for sequences of any lengths, a universal language model may have the ability of capturing semantics of free text and facilitating various applications that are highly dependent on the quality of language representation. In this subsection, we introduce a retrieval-based Natural Language Generation (NLG) task. 
The task is to generate articles based on manually created templates. Concretely, the goal is to retrieve one paragraph at a time from the corpus which best describes a certain sentence from the template and then combine the retrieved paragraphs into a complete passage. The main difficulty of this task lies in the need to compare semantics of sentence-level queries (usually contain only a few words) and paragraph-level documents (often consist of multiple sentences).

We use articles collected by our partners in a media company as our corpus. Each article is split into several paragraphs and each document contains one paragraph. The corpus has a total of 656k documents and cover a wide range of domains, including news, stories and daily scenarios. In addition, we have a collection of manually created templates in terms of 7 main categories, as shown in Table \ref{nlg-statistics}. Each template $T = \{s_1, s_2, \dots, s_N\}$ provides an outline of an article and contains up to $N=6$ sentences. Each sentence $s_i$ describes a particular aspect of the topic. 

The problem is solved in two steps. First, an index for all the documents is built using BM25. For each query, it will return a set of candidate documents that are related to the topic. Second, we use representation models to re-rank the top 100 candidates: each query-document pair $(\mathbf q, \mathbf d)$ is mapped to a score $f(\mathbf q, \mathbf d)$, where the scoring function $f$ is based on cosine similarity. Quality of the generated passages was assessed by two native Chinese speakers, who were asked to examine whether the retrieved paragraphs were ``relevant" to the topic and ``conveyed the meaning" of the given sentence.

\section{Implementation}
\subsection{Data Processing}
We download the English and Chinese Wikipedia Corpus\footnote{https://dumps.wikimedia.org} and pre-process with \texttt{process\_wiki.py}\footnote{https://github.com/panyang/Wikipedia\_Word2vec/blob/master\\/v1/process\_wiki.py}, which extracts text from xml files. Then for the Chinese corpus, we convert the data into simplified characters using \texttt{OpenCC}. In order to extract high-quality \textit{n}-grams, we remove punctuation marks and characters in other languages based on regular expressions, and finally get an English corpus of 2,266M words and a Chinese corpus of 380M characters. 

We calculate PMI scores of all \textit{n}-grams with a maximum length of $N=10$ for each document instead of the entire corpus considering that different documents usually describe different topics. We manually evaluate the extracted \textit{n}-grams and find nearly 50\% of the top 2000 \textit{n}-grams contain 3 $\sim$ 4 words (characters for Chinese), and only less than 0.5\% \textit{n}-grams are longer than 7. Although a larger \textit{n}-gram vocabulary can cover longer \textit{n}-grams, it will cause too many meaningless \textit{n}-grams. Therefore, for both English and Chinese corpus, we empirically retain the top 3000 \textit{n}-grams for each document, resulting in vocabularies of \textit{n}-grams with average lengths of 4.6 and 4.5, respectively. Finally, for English, we randomly sample 10M sentences rather than use the entire corpus to reduce training time.

\subsection{Pre-training}
As in BERT, sentence pairs are packed into a single sequence and the special \texttt{[CLS]} token is used for sentence-level predicting. While in accordance with Joshi et al. (2020) \cite{spanbert}, we find that single sentence training is better than the original sentence pair scenario. Thus in our experiments, the input is a continuous sequence with a maximum length of 512.

\begin{table*}[t]
\caption{Questions and candidates from the sentence-level ``positive-negative" analogy dataset and similarity scores for each candidate sentence computed by GloVe, USE and BURT-base. The correct sentences are in bold.}
  \label{pos-neg_example}
  \centering
  \begin{tabular}{lllllll}\toprule
    \multicolumn{7}{l}{\textit{Barton's inquiry was reasonable }:\textit{ Barton's inquiry was not reasonable }::\textit{ Changing the sign of numbers is an efficient algorithm}}\\
    
    \multicolumn{3}{l}\textit{\textbf{changing the sign of numbers is an inefficient algorithm}} & & GloVe: 0.96 & USE: 0.89\ & BURT-base: \textbf{0.97} \\
    
    \multicolumn{3}{l}\textit{changing the sign of numbers is not an inefficient algorithm} && GloVe: \textbf{0.97}&USE: \textbf{0.90}&BURT-base: 0.96  \\\midrule
    
    \multicolumn{7}{l}{\textit{Members are aware of their political work }:\textit{ Members are not aware of their political work }::\textit{ This ant is a known species}}\\
    
    \multicolumn{2}{l}\textit{\textbf{This ant is an unknown species}}  && &GloVe:0.94&USE:\textbf{0.87}&BURT-base: \textbf{0.96}\\
    
    \multicolumn{2}{l}\textit{This ant is not an unknown species} && &GloVe: \textbf{0.95}&USE:0.82 &BURT-base: 0.95  \\\bottomrule
  \end{tabular}
\end{table*}

\begin{table*}[t]
    \caption{Performance of different models on universal analogy datasets. Mean-pooling is applied to Transformer-based models to obtain fixed-length embeddings. The last column shows the average accuracy of word, phrase and sentence analogy tasks.}
  \label{analogy_acc}
  \centering
  \renewcommand{\multirowsetup}{\centering}
  \begin{tabular}{l|ccc|ccc|ccc|c}\toprule
  \multirow{2}{*}{\textbf{Models}}
  & \multicolumn{3}{c|}{\textbf{Word}} & \multicolumn{3}{c|}{\textbf{Phrase}} &
  \multicolumn{3}{c|}{\textbf{Sentence}}&\multirow{2}{*}{\textbf{Avg.}} \\
  \cline{2-10} \rule{0pt}{12pt}
    & semantic & syntactic & Avg. &semantic & syntactic & Avg. & semantic & syntactic & Avg. & \\\midrule
    
    GloVe \cite{glove} & 82.6 & 78.0	& 80.3	& \;\;0.0	& 40.9	& 20.5	& \;\;0.2 &	39.8 &	20.0 &	40.3  \\
    
    InferSent \cite{infersent}& 68.8 &	88.7 &	78.8&	\;\;0.0 &	54.1 &	27.0 &	\;\;0.0 &	50.8 &	25.4 &	43.7\\
    GenSen \cite{gensen}& 44.5 &	84.4 &	64.5 &	\;\;0.0 &	54.4 &	27.2 &	\;\;0.0 &	44.9 &	22.4 &	38.0 \\
    USE \cite{use}& 73.0 &	83.1 &	78.0 &	\;\;1.8 &	63.1 &	32.5 &	\;\;0.6 &	44.1 &	22.4 &	44.3 \\
    LASER \cite{laser} &26.9 & 78.2&52.6 &\;\;0.0 &63.3 &31.7 & \;\;1.6 & 55.4 & 28.5 &37.6 \\ 
    
    ALBERT-base \cite{albert}& 32.2	& 43.1	& 37.7	& \;\;0.0	& 56.4	& 28.2 &	\;\;0.0	& 59.2	& 29.6	& 31.8 \\
    ALBERT-xxlarge & 32.1	&37.5	&34.8	&\;\;0.9	&50.5	&25.7	&\;\;0.3	&50.3	&25.3	&28.6\\
    RoBERTa-base \cite{roberta}&  28.6	&50.5	&39.5	&\;\;0.0	&46.1	&23.0&	\;\;0.1&	63.6&	31.8&	31.5\\
    RoBERTa-large & 34.2&	55.9&	45.0&	\;\;0.2&	50.6&	25.4	&\;\;0.9&	50.9&	25.9&	32.1\\
    XLNet-base \cite{xlnet}&  23.2	 & 49.1 & 	36.1 & 	\;\;1.9 & 	65.6 & 	33.8 & 	\;\;0.8 & 	63.5	 & 32.2 & 	34.0\\
    XLNet-large & 23.4&	42.0&	32.7&	\;\;4.7	&53.5&	29.1&	\;\;5.6&	48.4&	27.0&	29.6\\\midrule
    
    BERT-base \cite{bert}& 51.3 &	60.2 &	55.8 &\;\;0.3 & 	69.3 &34.8	 &\;\;0.1 &	68.3 &34.2	&41.6	 \\
    MLM & 62.9 &61.1 &62.0&\;\;2.7&59.8&31.3& \;\;0.2&61.8& 31.0&41.4 \\
    Span & 63.5&58.9&61.2 &\;\;1.9&68.9&35.4&\;\;0.1&63.1&31.6& 42.7\\
    \textbf{BURT-base} &71.1 & 74.4 &72.8 & \;\;1.7 & 69.1 &35.4 & \;\;0.6 & 63.4&32.0 & 46.7  \\\midrule

    BERT-large & 49.7 &	46.6 &	48.2 &\;\;0.1 & 67.4&33.9 &\;\;0.5	 &61.2 	&30.9& 37.7\\
    MLM &65.0& 50.7&57.9 &\;\;1.0&63.4&32.2&\;\;0.5&56.7&28.6&39.6 \\
    Span &66.5&54.1&60.3&\;\;2.7&64.2&33.5& \;\;0.7&58.4&29.6&40.5\\
     \textbf{BURT-large}&84.7 &74.0 &79.4 &\;\;4.9& 58.6&31.8 &\;\;1.0 &52.4 &26.7&46.0  \\\midrule
    
    SBERT-base \cite{sbert}& 71.2	&73.7&	72.4&	41.8&	63.6&	52.7&	23.2&	58.7&	40.9&	55.3\\ 
    \textbf{SBURT-base} &82.8 &77.5  &80.2 & 33.2 &70.5 &51.9 &30.8  & 69.1&\textbf{50.0} & \textbf{60.7} \\
    
    SBERT-large& 72.5	&74.2	&73.3&	57.8&	55.0&	\textbf{56.4}&	18.4&	52.4&	35.4	&55.0\\ 
    \textbf{SBURT-large} &84.4 & 76.6 &\textbf{80.5}& 34.7&50.1&42.4&5.7&53.1&29.4&50.8 \\
    \bottomrule
  \end{tabular}
\end{table*}

Instead of training from scratch, we initialize both English and Chinese models with the officially released checkpoints (\texttt{bert-base-uncased}, \texttt{bert-large-uncased}, \texttt{bert-base-chinse}) and BERT-wwm-ext, which is trained from the Chinese BERT using whole word masking on extended data \cite{bert-wwm}. Base models are comprised of 12 Transformer layers, 12 heads, 768 dimensional hidden states and 110M parameters in total. The English BERT-large has 24 Transformer layers, 16 heads, 1024 dimensional hidden states and 340M parameters in total. We use Adam optimizer \cite{adam} with initial learning rate of 5e-5 and linear warmup over the first 10\% of the training steps. Batch size is set to 16 and dropout rate is 0.1. Each model is trained for one epoch.

\subsection{Fine-tuning}
Following BERT, in the fine-tuning procedure, pairs of sentences are concatenated into a single sequence with a special token \texttt{[SEP]} in between. For both single sentence and sentence pair tasks, the hidden state of the first token \texttt{[CLS]} is used for softmax classification. We use the same sets of hyperparameters for all the evaluated models. All experiments on the GLUE benchmark are ran 
with a total train batch sizes between 8 and 64 and learning rates of 3e-5 for 3 epochs. For tasks from the CLUE benchmark, we set batch sizes to 8 and 16, learning rates between 1e-5 and 3e-5, and train 50 epochs on WSC and 2$\sim$5 epochs on the rest tasks.

\subsection{Downstream-task Models}
On GLUE and CLUE, we compare our model with three variants: pre-trained models (Chinese BERT/BERT-wwm-ext, English BERT-base/BERT-large), models trained with the same number of additional steps as our model (MLM), and models trained using random span masking with the same number of additional steps as our model (Span). For the Span model, we simply replace our \textit{n}-gram module with the masking strategy as proposed by \cite{spanbert}, where the sampling probability of span length $l$ is based on a geometric distribution $l \sim Geo(p)$. We follow the parameter setting that $p=0.2$ and maximum span length $l_{max}=10$.

We also evaluate the aforementioned models on our universal analogy task. Baseline models include Bag-of-words (BoW) model from pre-trained word embeddings: GloVe, sentence embedding models: InferSent, GenSen, USE and LASER, pre-trained contextualized language models: BERT, ALBERT, RoBERTa and XLNet. To derive semantically meaningful embeddings, we fine-tune BERT and our model on the Stanford Natural Language Inference (SNLI) \cite{snli} and the Multi-Genre NLI Corpus \cite{multinli} using a Siamese structure following Reimers and Gurevych (2019) \cite{sbert}.

\begin{table*}[t]{}
\caption{Examples of the retrieved paragraphs and corresponding comments from the judges. ``B"-BM25, ``L"-LASER, ``S"-Span, ``U"-BURT.}
  \label{nlg-example}
  \centering
  \begin{tabular}{p{15cm}}\toprule
  
  \textbf{Query}: \begin{CJK}{UTF8}{gbsn}
        端午节的\textbf{由来}
    \end{CJK}(\textit{The \textbf{Origin} of the Dragon Boat Festival}) 
    \\\midrule
    
    \begin{CJK}{UTF8}{gbsn}
    $B$: 一个中学的高级教师陈老师生动地解读端午节的\textbf{由来}，诵读爱好者进行原创诗歌作品朗诵，深深打动了在场的观众...
    \end{CJK}
    (\textit{Mr. Chen, senior teacher at a middle School, vividly introduced the \textbf{origin} of the Dragon Boat Festival and people are reciting original poems, which deeply moved the audience...}) \\\midrule
    
    \begin{CJK}{UTF8}{gbsn}
        $L$: 今天是端午小长假第一天...当天上午，在车厢内满目挂有与端午节相关的民俗故事及有关诗词的文字...
    \end{CJK}(\textit{Today is the first day of the Dragon Boat Festival holiday...There are folk stories and poems posted in the carriage...}) \\\midrule
    
     \begin{CJK}{UTF8}{gbsn}
        $S, U$: ...端午节又称端阳节、龙舟节、浴兰节，是中华民族的传统节日。端午节形成于先秦，发展于汉末魏晋，兴盛于唐...
    \end{CJK}(\textit{...Dragon Boat Festival, also known as Duanyang Festival, Longzhou Festival and Yulan Festival is a traditional festival of the Chinese nation. It is formed in the Pre-Qin Dynasty, developed in the late Han and Wei-Jin, and prospered in the Tang...}) \\\midrule

  \textbf{Comments}: $B$ and $L$ is related to the topic but does not convey the meaning of the query.\\ \midrule\midrule

    \textbf{Query}: \begin{CJK}{UTF8}{gbsn}
        狗的\textbf{喂养}知识
    \end{CJK}(\textit{Dog \textbf{Feeding} Tips}) \\\midrule
    
    \begin{CJK}{UTF8}{gbsn}
    $B$: ...创建一个“比特狗”账户，并支付99元领养一只“比特狗”。然后购买\textbf{喂养}套餐喂养“比特狗”，“比特狗”就能通过每天挖矿产生BTGS虚拟货币。
    \end{CJK}(\textit{...First create a ``Bitdog" account and pay 99 yuan to adopt a ``Bitdog". Then buy a package to \textbf{feed} the ``Bitdog", which can generate virtual currency BTGS through daily mining.}) \\\midrule
    
    \begin{CJK}{UTF8}{gbsn}
      $L$: 要养成定时定量喂食的好习惯，帮助狗狗更好的消化和吸收，同时也要选择些低盐健康的狗粮...
    \end{CJK}(\textit{It is necessary to feed your dog regularly and quantitatively to help them digest and absorb better. Meanwhile, choose some low-salt and healthy food...}) \\\midrule
    
    \begin{CJK}{UTF8}{gbsn}
      $S$: 泰迪犬容易褪色是受到基因和护理不当的影响，其次是饮食太咸...一定要注意正确护理，定期洗澡，要给泰迪低盐营养的优质狗粮...
    \end{CJK}(\textit{Teddy bear dog's hair is easy to fade because of its genes and improper care. It is also caused by salty diet... So we must take good care of them, such as taking a bath regularly, and preparing dog food with low salt...}) \\\midrule
    
    \begin{CJK}{UTF8}{gbsn}
      $U$: 还可以一周自制一次狗粮给狗狗喂食，就是买些肉类，蔬菜，自己动手做。偶尔吃吃自制狗粮也能增加狗狗的营养，和丰富狗狗的口味。日常的话，建议选择些适口性强的狗粮，有助磨牙，防止口腔疾病。
    \end{CJK}(\textit{You can also make dog food once a week, such as meats and vegetables. Occasionally eating homemade dog food can also supplement nutrition and enrich the taste. In daily life, it is recommended to choose some palatable dog food to help their teeth grinding and prevent oral diseases.}) \\ \midrule
    
     \textbf{Comments}: $B$ is not a relevant paragraph. $S$ is relevant to the topic but is inaccurate.\\ \bottomrule
     
  \end{tabular}
\end{table*}

\begin{table}[t]
 \caption{Comparison of models performance on the FAQ dataset.}
  \label{faq-results}
  \centering
\begin{tabular}{l|rr}\arrayrulecolor{black} \toprule
    Method  & Acc. &MRR\\\midrule
    TF-IDF  &	73.7 & 0.813\\
    BM25	&72.1 & 0.802\\
    LASER & 79.9 & 0.856 \\\midrule
    BERT&	76.8&0.831 \\
    MLM & 78.3 & 0.843 \\
    Span & 78.6&0.846 \\
  \textbf{BURT} &	\textbf{82.2} &\textbf{0.872}\\\midrule
    BERT-wwm-ext&76.7   &0.834  \\
    MLM & 76.7& 0.834\\
    Span &79.3 & 0.856\\
    \textbf{BURT-wwm-ext}& 80.7     &0.863 \\\bottomrule
  \end{tabular}
\end{table}

\begin{table}[t]
  \caption{Results on NLG according to human judgment. ``R" and ``CM" represent the percentage of paragraphs that are ``relevant" and ``convey the meaning", respectively. }
  \label{nlg-results}
  \centering
\begin{tabular}{l|rrrrrr}\arrayrulecolor{black} \toprule
    \textbf{R}  & BM25  &LASER & BERT & MLM & SPAN& BURT\\\midrule
    Judge1	&60.3& 63.9 & 65.9&65.0&69.3&71.8\\
    Judge2 &61.8&61.6 & 67.3&67.5&71.5&71.0\\
    Avg. &  61.1&62.8 & 66.6&66.3&70.4&71.4\\\midrule
    \textbf{CM}  & BM25  &LASER & BERT & MLM & SPAN& BURT\\\midrule
    Judge1	& 43.5&42.5&48.5&46.1&51.6&54.2\\
    Judge2 & 41.2&38.4&47.8&45.5&53.9&56.5\\
    Avg. &42.4  &40.5&48.2&45.8&52.8&55.4\\\bottomrule
  \end{tabular}
\end{table}

For FAQ and NLG, we compare our models with statistical methods such as TF-IDF and BM25, a sentence representation model LASER \cite{laser}, the pre-trained BERT/BERT-wwm-ext and models trained with additional steps (MLM, Span). We observe that further training on Chinese SNLI and MNLI datasets underperforms BERT on the FAQ dataset. Therefore, we only consider pre-trained models for these two tasks.

\section{Experiments}
\subsection{General Language Understanding}
Table \ref{clue-results} and {} show the results on the GLUE and CLUE benchmarks, where we find that training BERT with additional MLM steps can hardly bring any improvement except for the WSC task. In Chinese, the Span model is effective on WSC but is comparable to BERT on other tasks. BERT-wwm-ext is better than our model on classification tasks involving pairs of short sentences such as AFQMC and OCNLI, which may be due to its relative powerful capability of modeling short sequences. Overall, both BURT and BURT-wwm-ext outperform the baseline models on 4 out of 6 tasks with considerable improvement, which sheds light on their effectiveness of modeling sequences of different lengths. The most significant improvement is observed on WSC (3.5\% over the updated BERT-wwm-ext and 0.7\% over the Span model), where the model is trained to determine whether the given two spans refer to the same entity in the text. We conjecture that the model benefits from learning to predict meaningful spans in the pre-training stage, so it is better at capturing the meanings of spans in the text. In English, our approach also improves the performance of BERT on various tasks from the GLUE benchmark, indicating that our proposed PMI-based masking method is general and independent with language settings.

\subsection{Universal Analogy}
Results on analogy tasks are reported in Table \ref{analogy_acc}. Generally, semantic analogies are more challenging than the syntactic ones and higher-level relationships between sequences are more difficult to capture, which is observed in almost all the evaluated models. On word analogy tasks, all well pre-trained language models like BERT, ALBERT, RoBERTa and XLNet hardly exhibit arithmetic characteristics and increasing the model size usually leads to a decrease in accuracy. However, our method of pre-training using \textit{n}-grams extracted by the PMI algorithm significantly improves the performance on word analogies compared with BERT, obtaining 72.8\% (BURT-base) and 79.4\% (BURT-large) accuracy, respectively. Further training BURT-large on SNLI and MNLI results in the highest accuracy (80.5\%).

Despite the leading performance on word-level analogy datasets of GloVe, InferSent and USE, they do not generalize well on higher level analogy tasks. We conjecture their poor performance is caused by synonyms and paraphrases in sentences which lead the model to produce lower similarity scores to the correct answers. In contrast, Transformer-based models are more advantageous in representing higher-level sequences and are good at identifying paraphrases and capturing relationships between sentences even if they have less lexical overlap. Moreover, fine-tuning pre-trained models achieves considerable improvements on high-level semantic analogies. Overall, SBURT-base achieves the highest average accuracy (60.7\%). 

Examples from the Negation subset are shown in Table \ref{pos-neg_example}. Notice that the word ``\textit{not}" does not explicitly appear in the correct answers. Instead, ``\textit{inefficient}" and ``\textit{unaware}" are indicators of negation. As expected, BOW will give a higher similarity score for the sentence that contain both ``\textit{not}" and ``\textit{inefficient}" because the word-level information is simply added and subtracted despite the context. By contrast, contextualized models like BURT capture the meanings and relationships of words within the sequence in a comprehensive way, indicating that it has indeed learned universal representations across different linguistic units.

\begin{table}[t]
\caption{Annotation of phrases and sentences in Figure \ref{distance}.}
  \label{explanation}
  \begin{tabular}{|ll|}\hline
   
    p\_man: \textit{employed by the \underline{man}} & p\_woman: \textit{hired by the \underline{woman}}\\
    
    p\_king: \textit{employed by the \underline{king}} & p\_queen: \textit{hired by the \underline{queen}}\\
   
  p\_dad: \textit{employed by his \underline{dad}} & p\_mom: \textit{hired by his \underline{mom}}\\\hline
   
  \multicolumn{2}{|l|}{s\_man: \textit{He was employed by the \underline{man} when he was 28.}} \\
    \multicolumn{2}{|l|}{s\_woman: \textit{He was hired by the \underline{woman} at age 28.}} \\
    \multicolumn{2}{|l|}{s\_king: \textit{He was employed by the \underline{king} when he was 28.}} \\
    \multicolumn{2}{|l|}{s\_queen: \textit{He was hired by the \underline{queen} at age 28.}} \\
    \multicolumn{2}{|l|}{s\_dad: \textit{He was employed by his \underline{dad} when he was 28.}} \\
  \multicolumn{2}{|l|}{s\_mom: \textit{He was hired by his \underline{mom} at age 28.}} \\\hline
  \end{tabular}
\end{table}

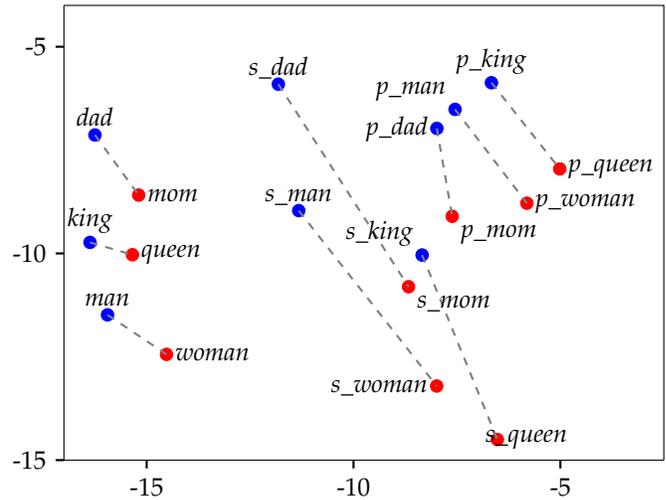
\begin{figure}[t]
\centering
\begin{tikzpicture} [xscale=0.55, yscale=0.55]
\draw [-] (-17,-4) -- (-17,-15) -- (-2.5,-15);
\draw [-] (-17,-4) -- (-2.5,-4) -- (-2.5,-15);
\draw [thick] (-15,-15.2) node[below]{-15} -- (-15,-15);
\draw [thick] (-10,-15.2) node[below]{-10} -- (-10,-15);
\draw [thick] (-5,-15.2) node[below]{-5} -- (-5,-15);
\draw [thick] (-17.2,-15) node[left]{-15} -- (-17,-15);
\draw [thick] (-17.2,-10) node[left]{-10} -- (-17,-10);
\draw [thick] (-17.2,-5) node[left]{-5} -- (-17,-5);

\draw [fill,blue] ( -23.36894  +7  , 0.26676014-10) circle [radius=0.15];
\draw [fill,blue] (-23.250492 +7 ,  2.8655763-10) circle [radius=0.15];
\draw [fill,blue] (-22.94649 +7,  -1.486137-10) circle [radius=0.15];
\draw [fill,blue] (-4.666675 -2,-4.870993-1) circle [radius=0.15];
\draw [fill,blue] (-5.9862328 -2,-5.9751067-1) circle [radius=0.15];
\draw [fill,blue] (-5.5463557 -2,-5.515645 -1) circle [radius=0.15];
\draw [fill,blue] (-4.3424816 -4,-13.038456+3) circle [radius=0.15];
\draw [fill,blue] (-7.8167977 -4,-8.905666+3) circle [radius=0.15];
\draw [fill,blue] (-7.324448 -4,-11.963703+3) circle [radius=0.15];

\draw [fill,red] ( -22.343668 +7,   -0.03237649-10) circle [radius=0.15];
\draw [fill,red] (-22.192139  +7,  1.4110558-10) circle [radius=0.15];
\draw [fill,red] (-21.520405 +7  ,-2.4419914-10) circle [radius=0.15];
\draw [fill,red] (-3.0148447 -2,-6.9562917-1) circle [radius=0.15];
\draw [fill,red] (-5.615085 -2, -8.1017447-1) circle [radius=0.15];
\draw [fill,red] (-3.810241 -2, -7.7853274-1) circle [radius=0.15];
\draw [fill,red] (-3.0234008 -3.5,-17.500946+3) circle [radius=0.15];
\draw [fill,red] (-4.1695065-4.5 ,-13.805784+3) circle [radius=0.15];
\draw [fill,red] (-3.9898653-4,-16.207256+3) circle [radius=0.15];

\draw [dashed, gray, thick] (-23.36894  +7  , 0.26676014-10) -- (-22.343668 +7,   -0.03237649-10);
\draw [dashed, gray, thick] (-23.250492 +7 ,  2.8655763-10) -- (-22.192139  +7,  1.4110558-10);
\draw [dashed, gray, thick] (-22.94649 +7,  -1.486137-10) -- (-21.520405 +7  ,-2.4419914-10);
\draw [dashed, gray, thick] (-4.666675 -2,-4.870993-1) -- (-3.0148447 -2,-6.9562917-1);
\draw [dashed, gray, thick] (-5.9862328 -2,-5.9751067-1) -- (-5.615085 -2, -8.1017447-1);
\draw [dashed, gray, thick] (-5.5463557 -2,-5.515645 -1) -- (-3.810241 -2, -7.7853274-1);
\draw [dashed, gray, thick] (-4.3424816 -4,-13.038456+3) -- (-3.0234008 -3.5,-17.500946+3);
\draw [dashed, gray, thick] (-7.8167977 -4,-8.905666+3) -- (-4.1695065-4.5 ,-13.805784+3);
\draw [dashed, gray, thick] (-7.324448 -4,-11.963703+3) -- (-3.9898653-4,-16.207256+3);

\node [above,font=\itshape] at (-23.36894  +7  , 0.26676014-10) {king};
\node [right,font=\itshape] at (-22.343668 +7,   -0.03237649-10) {queen};
\node [above,font=\itshape] at (-23.250492 +7 ,  2.8655763-10) {dad};
\node [right,font=\itshape] at (-22.192139  +7,  1.4110558-10) {mom};
\node [above,font=\itshape] at (-22.94649 +7,  -1.486137-10) {man};
\node [right,font=\itshape] at (-21.520405 +7  ,-2.4419914-10) {woman};

\node [above,font=\itshape] at (-4.666675 -2,-4.870993-1) {p\_king};
\node [right,font=\itshape] at (-3.0148447 -2,-6.9562917-1) {p\_queen};
\node [left,font=\itshape] at (-5.9862328 -2,-5.9751067-1) {p\_dad};
\node [below right,font=\itshape] at (-5.615085 -2, -8.1017447-1) {p\_mom};
\node [above left,font=\itshape] at (-5.5463557 -2,-5.515645 -1) {p\_man};
\node [right,font=\itshape] at (-3.810241 -2, -7.7853274-1) {p\_woman};

\node [above left,font=\itshape] at (-4.3424816 -4,-13.038456+3) {s\_king};
\node [right,font=\itshape] at (-3.0234008 -4,-17.500946+3) {s\_queen};
\node [above,font=\itshape] at (-7.8167977 -4,-8.905666+3) {s\_dad};
\node [below right,font=\itshape] at (-4.1695065-4.5 ,-13.805784+3) {s\_mom};
\node [above,font=\itshape] at (-7.324448 -4,-11.963703+3) {s\_man};
\node [left,font=\itshape] at (-3.9898653-4,-16.207256+3) {s\_woman};
\end{tikzpicture}
\caption{Two-dimensional PCA projection of the vectors representing ``male" and ``female" generated by BURT. Pairs are connected by dashed lines. Points in the figure as explained in detail in Table \ref{explanation}.}
\label{distance}
\end{figure}

\begin{figure}[t]
\centering
\includegraphics[width=0.45\textwidth]{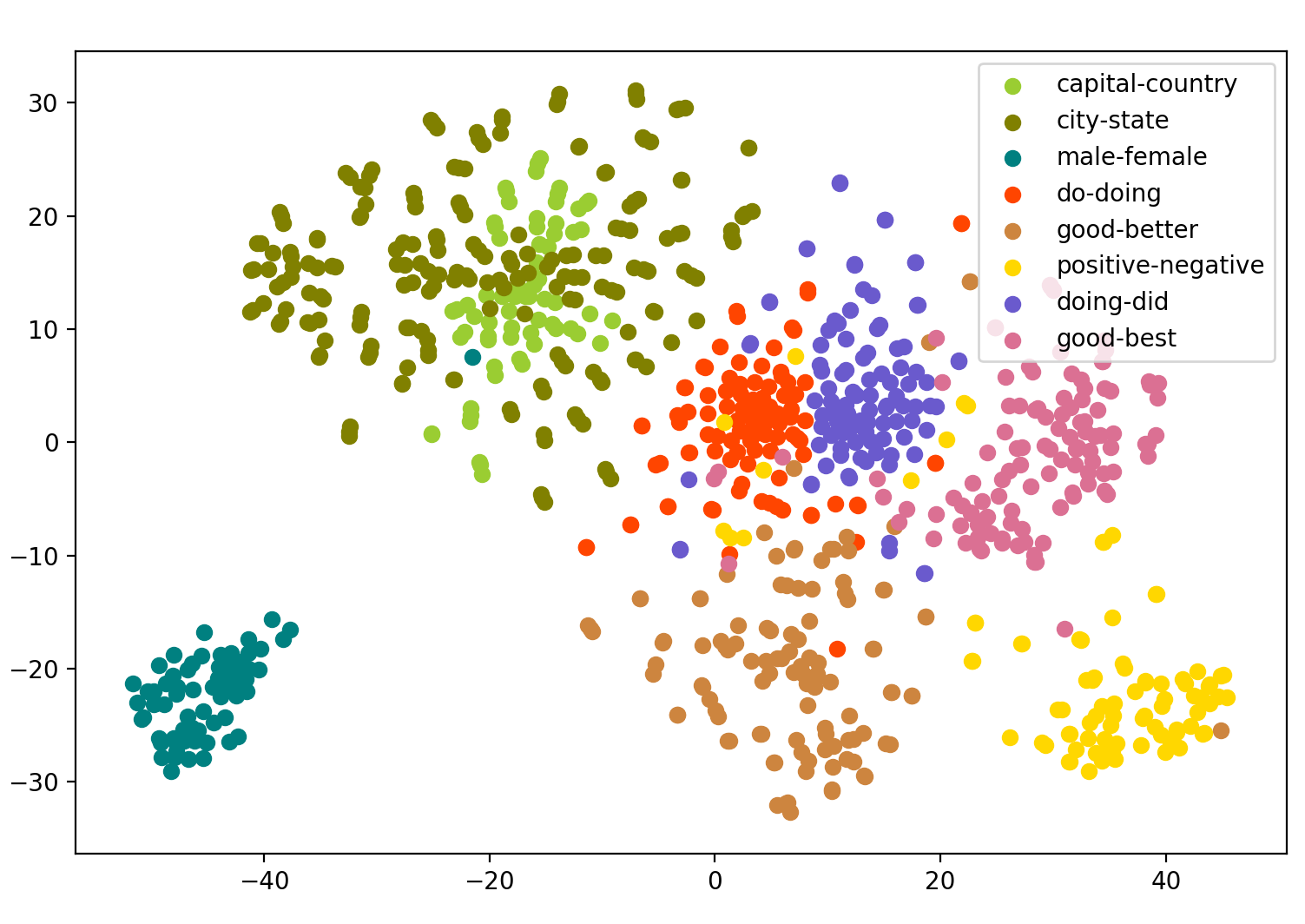}
\caption{t-SNE projection of patterns.}
\label{cluster}
\end{figure}

\subsection{Retrieval-based FAQ}
Results are reported in Table \ref{faq-results}. As we can see, LASER and all pre-trained language models significantly outperform TF-IDF and BM25, indicating the superiority of embedding-based models over statistical methods. Besides, the continued BERT training is often beneficial. Among all the evaluated models, our BURT yields the highest accuracy (82.2\%) and MRR (0.872). BURT-wwm-ext achieves a slightly lower accuracy (80.7\%) compared with BURT but it still exceeds its baselines by 4.0\% (MLM) and 1.4\% (Span), respectively.

\subsection{Natural Language Generation}
Results are summarized in Table \ref{nlg-results}. Although nearly 62\% of the paragraphs retrieved by BM25 are relevant to the topic, only two-thirds of them actually convey the original meaning of the template. Despite LASER's comparable performance to BURT on FAQ, it is less effective when different granular linguistic units are involved at the same time. Re-ranking using BURT substantially improves the quality of the generated paragraphs. We show examples retrieved by BM25 , LASER, the Span model and BURT in Table \ref{nlg-example}, denoted by $B$, $L$, $S$ and $U$, respectively. BM25 tends to favor paragraphs that contain the keywords even though the paragraph conveys a different meaning, while BURT selects accurate answers according to semantic meanings of queries and documents. 

\section{Visualization}
\subsection{Single Pattern}
Mikolov et al. (2013) \cite{phrase2vec} use PCA to project word embeddings into a two-dimensional space to visualize a single pattern captured by the Word2Vec model, while in this work we consider embeddings for different granular linguistic units. All pairs in Figure \ref{distance} belong to the ``male-female" category and subtracting the two vectors results in roughly the same direction. 

\subsection{Clustering}
Given that embeddings of sequences with the same kind of relationship will exhibit the same pattern in the vector space, we obtain the difference between pairs of embeddings for words, phrases and sentences from different categories and visualize them by t-SNE. Figure \ref{cluster} shows that by subtracting two vectors, pairs that belong to the same category automatically fall into the same cluster. Only the pairs from ``capital-country" and ``city-state" cannot be totally distinguished, which is reasonable because they all describe the relationship between geographical entities.

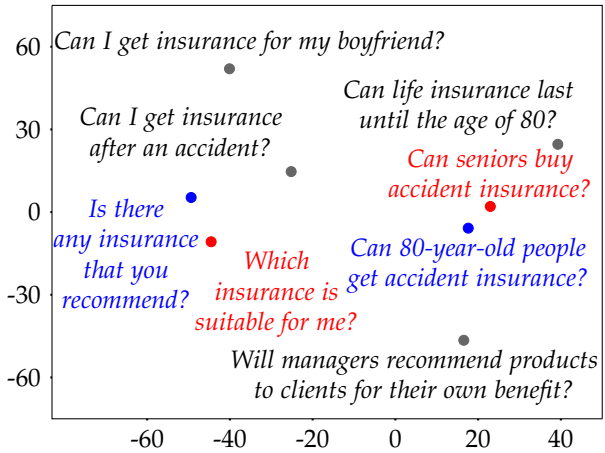
\begin{figure}[t]
\centering
\begin{tikzpicture} [xscale=0.11, yscale=0.11]
\draw [-] (-41.5,25) -- (-41.5,-25) -- (25,-25);
\draw [-] (-41.5,25) -- (25,25) -- (25,-25);
\draw [thick] (-30,-25.2) node[below]{-60} -- (-30,-25);
\draw [thick] (-20,-25.2) node[below]{-40} -- (-20,-25);
\draw [thick] (-10,-25.2) node[below]{-20} -- (-10,-25);
\draw [thick] (0,-25.2) node[below]{0} -- (0,-25);
\draw [thick] (10,-25.2) node[below]{20} -- (10,-25);
\draw [thick] (20,-25.2) node[below]{40} -- (20,-25);
\draw [thick] (-41.7,-20) node[left]{-60} -- (-41.5,-20);
\draw [thick] (-41.7,-10) node[left]{-30} -- (-41.5,-10);
\draw [thick] (-41.7,0) node[left]{0} -- (-41.5,0);
\draw [thick] (-41.7,10) node[left]{30} -- (-41.5,10);
\draw [thick] (-41.7,20) node[left]{60} -- (-41.5,20);

\draw [fill,blue] (21.611923/2 -2,    0.16701147/3-2) circle [radius=0.6];
\draw [fill,red] (22.951221/2,   2.0454462/3 ) circle [radius=0.6];
\draw [fill,blue] (-24.68343,    5.293025/3) circle [radius=0.6]; 
\draw [fill,red] (-24.267378+2,    4.2287335/3 -5) circle [radius=0.6]; 
\draw [fill,black!60] (16.53517/2, -46.51374/3) circle [radius=0.6];  
\draw [fill,black!60] (39.26717 /2, 24.598452/3) circle [radius=0.6]; 
\draw [fill,black!60] (-20.049093 , 51.999535/3) circle [radius=0.6];
\draw [fill,black!60] (-12.593758,  14.688758/3) circle [radius=0.6]; 

\node [below,align=center,blue, font=\itshape] at (21.611923/2-2,    0.16701147/3-2) {Can 80-year-old people\\get accident insurance?};
\node [above,align=center,red, font=\itshape] at (22.951221/2,   2.0454462/3 ) {Can seniors buy\\accident insurance?};
\node [below,align=center,black, font=\itshape] at (16.53517/2-6, -46.51374/3) {Will managers recommend products\\to clients for their own benefit?};
\node [above left,align=center,black, font=\itshape] at (39.26717 /2+3, 24.598452/3) {Can life insurance last\\until the age of 80?};

\node [below left,align=center,blue, font=\itshape] at (-24.68343+1+1,    5.293025/3+1) {Is there\\any insurance\\that you\\recommend?};
\node [below right,align=center,red, font=\itshape] at (-24.267378-1,    4.2287335/3 -5) {Which\\insurance is\\suitable for me?};
\node [above,align=center,black, font=\itshape] at (-20.049093+2+0.5 , 51.999535/3+0.5) {Can I get insurance for my boyfriend?};
\node [above left,align=center,black, font=\itshape] at (-12.593758,  14.688758/3) {Can I get insurance\\after an accident?};
\end{tikzpicture}
\caption{t-SNE projection of BURT embeddings. Blue dots: queries, Red dots: sentences retrieved by BURT, Grey dots: sentences retrieved by TF-IDF and BM25. }
\label{faq_vector}
\end{figure}

\subsection{FAQ}
We show examples in Figure \ref{faq_vector} where BURT successfully retrieve the correct answer while TF-IDF and BM25 fail. Both sentences ``\textit{Can 80-year-old people get accident insurance?}" and ``\textit{Can life insurance last until the age of 80?}" contain the word ``\textit{80}", which is a possible reason why TF-IDF tends to believe they highly match with each other, ignoring that the two sentences are actually describing two different issues. In contrast, using vector-based representations, BURT considers ``\textit{seniors}" as a paraphrase of ``\textit{80-year-old people}". As depicted in Figure \ref{faq_vector}, queries are close to the correct responses and away from other sentences.

\section{Conclusion}
This paper formally introduces the task of universal representation learning and then presents a pre-trained language model for such a purpose to map different granular linguistic units into the same vector space where similar sequences have similar representations and enable unified vector operations among different language hierarchies. 

In detail, we focus on the less concentrated language representation, seeking to learn a uniform vector form across different linguistic unit hierarchies. Far apart from learning either word only or sentence only representation, our method extends BERT's masking and training objective to a more general level, which leverage information from sequences of different lengths in a comprehensive way and effectively learns a universal representation from words, phrases to sentences. 

Overall, our proposed BURT outperforms its baselines on a wide range of downstream tasks with regard to sequences of different lengths in both English and Chinese languages. We especially provide an universal analogy task, an insurance FAQ dataset and an NLG dataset for extensive evaluation, where our well-trained universal representation model holds the promise for demonstrating accurate vector arithmetic with regard to words, phrases and sentences and in real-world retrieval applications. 



\ifCLASSOPTIONcaptionsoff
  \newpage
\fi


\bibliographystyle{IEEEtran}
\bibliography{ref}

%

\begin{IEEEbiography}[{\includegraphics[width=1in,height=1.25in,clip,keepaspectratio]{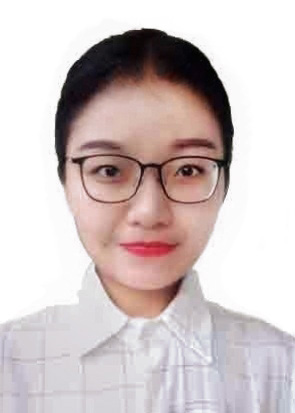}}]{Yian Li}
received her Bachelor's degree in Electronic and Information Engineering from Huazhong University of Science and Technology in 2019. She is working towards the M.S. degree in Department of Computer Science, Shanghai Jiao Tong University. Her research interests focuses on natural language processing, especially in language representation and pre-training language model.
\end{IEEEbiography}

\begin{IEEEbiography}[{\includegraphics[width=1in,height=1.25in,clip,keepaspectratio]{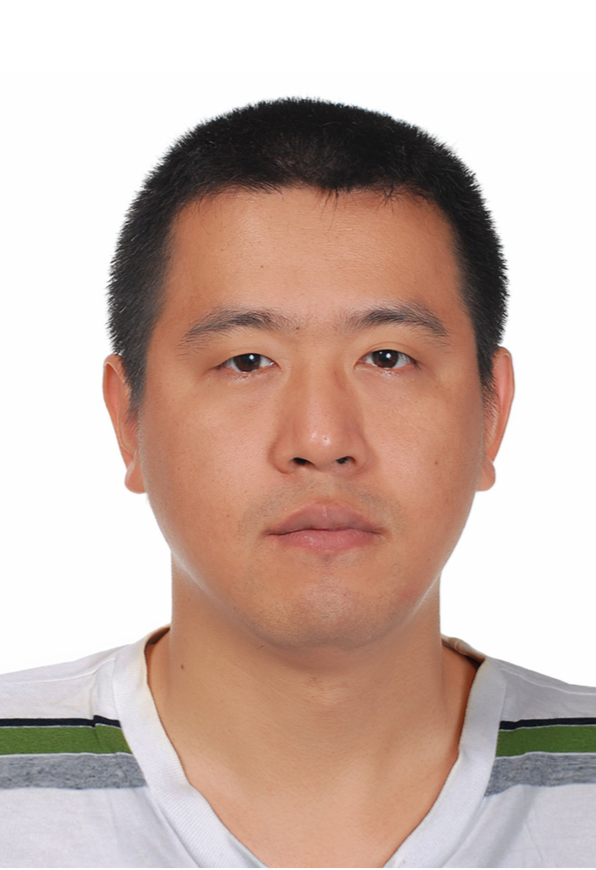}}]{Hai Zhao}
received the BEng degree in sensor and instrument engineering, and the MPhil degree in control theory and engineering from Yanshan University in 1999 and 2000, respectively, and the PhD degree in computer science from Shanghai Jiao Tong University, China in 2005. He is currently a full professor at department of computer science and engineering, Shanghai Jiao Tong University after he joined the university in 2009. He was a research fellow at the City University of Hong Kong from 2006 to 2009, a visiting scholar in Microsoft Research Asia in 2011, a visiting expert in NICT, Japan in 2012. He is an ACM professional member, and served as area co-chair in ACL 2017 on Tagging, Chunking, Syntax and Parsing, (senior) area chairs in ACL 2018, 2019 on Phonology, Morphology and Word Segmentation. His research interests include natural language processing and related machine learning, data mining and artificial intelligence.
\end{IEEEbiography}





\end{document}